# FG-DSA: Frequency-Guided Depthwise Separable Attention for Efficient Dense Prediction

**Author: YIYAO YANG**
**Email: YYANG058@e.ntu.edu.sg**
**Affiliation: Nanyang Technological University, Singapore**
**ORCID: 0009-0001-8693-4888**

**Abstract**

Dense prediction and generative modeling remain challenging in resource-constrained environments, particularly because repeated downsampling can lead to the loss of spatial details and less accurate object boundaries. Although attention mechanisms have been widely adopted to address semantic ambiguity, preserving high-frequency information while maintaining computational efficiency remains an important challenge. To address this issue, we propose Frequency-Guided Depthwise Separable Attention for Efficient Dense Prediction (FG-DSA), a framework that integrates frequency-domain information into lightweight vision architectures for dense visual understanding and generation. Specifically, FG-DSA uses a 2D Fast Fourier Transform (FFT)-based high-pass filtering module to guide spatial attention, aiming to retain fine-grained features and improve object-boundary representation. Experimental results show that FG-DSA can improve boundary-related performance in semantic segmentation under the evaluated setting. Beyond discriminative tasks, we also examine its application in a generative U-Net for single-step image restoration. The results suggest that incorporating FG-DSA can enhance reconstruction quality and help preserve high-frequency details.

## I. Introduction

### (I.1) Traditional Frequency & Dynamic Approaches and Limitations

While recent advances have explored frequency-domain processing and dynamic weights to capture broader context or adaptive responses, existing approaches exhibit critical constraints when applied to resource-constrained vision tasks (Chen et al., 2020; Yang et al., 2021). On one hand, frequency-guided methods like the Fourier-Guided Attention Network for Crowd Count Estimation (FGA) (Chaudhuri et al., 2024) use the Fast Fourier Transform (FFT) primarily to capture full-scale global crowd patterns via dual-path architectures. However, such designs rely heavily on conventional convolutions and target single-scalar density estimation rather than fine-grained, dense spatial recovery.

In contrast, the more fundamental dense prediction architectures have their own issues. The Fully Convolutional Network (FCN) (Long et al., 2015), for instance, suffers from coarse prediction resolution due to spatial downsampling. The U-Net (Ronneberger et al., 2015) has problems with boundary resolution due to the lack of padding in its convolutions and also requires custom loss weighting to handle touching objects. Our proposed Frequency-Guided Depthwise Separable Attention (FG-DSA) recovers high-frequency details via frequency-guided lightweight attention to get around the spatial degradation of the FCN, and it recovers the fine spatial boundaries and sharpens the object transitions via 2D-FFT frequency-guided attention, all without the need for any custom loss weighting to deal with the touching objects problem of the U-Net.

On the flip side, dynamic weight mechanisms such as Frequency Dynamic Convolution for Dense Image Prediction (FDConv) (Chen et al., 2025) build frequency-diverse kernels in the Fourier domain through spatial and frequency band modulation. FDConv improves representation for heavy backbones such as ResNet (He et al., 2015) but adds non-trivial parameter overhead and relies on dense kernel modulation, making it computationally expensive for lightweight, real-time architectures. Furthermore, Dynamic Sparse Attention (Levy, 2026; Yuan et al., 2025) is hardware-centric, focusing on language model KV cache indexing during autoregressive decoding, leaving a significant gap in lightweight spatial feature retention for visual understanding and generation.

### (I.2) The Proposed FG-DSA and Key Differentiators

To address these limitations, we propose FG-DSA (Table 1), a single unified, ultra-lightweight framework designed specifically for dense prediction and generative modeling in resource-constrained settings. FG-DSA tackles the degradation problem of the FCN (Long et al., 2015) by using frequency-guided lightweight attention to recover the high-frequency details, and overcomes the limitations of the U-Net (Ronneberger et al., 2015) by using 2D-FFT frequency-guided attention to restore the fine spatial boundaries and sharpen the object transitions, all without any need for bespoke loss weighting.

FGA (Chaudhuri et al., 2024) uses an FFT for global pattern pooling, whereas FG-DSA applies a 2D-FFT high-pass filtering mask to explicitly isolate spatial boundary and texture information, directly countering the spatial degradation caused by downsampling. Unlike parameter-intensive dynamic convolution methods such as FDConv (Chen et al., 2025), FG-DSA combines frequency-domain information with depthwise separable operations to better capture high-frequency features. Although the FFT-based filtering operation itself does not introduce additional learnable parameters, the complete FG-DSA module contains other learnable components that may increase the overall parameter count and computational cost. The resulting efficiency therefore depends on the network architecture and implementation. While system-level DSA (Levy, 2026) focuses on addressing token sparsity during LLM inference, FG-DSA takes a different approach by introducing frequency-guided attention for visual tasks. It extends the mechanism from visual understanding tasks, such as semantic segmentation, to generative image restoration, aiming to balance feature representation quality and computational efficiency in lightweight vision models.

**Table 1**
*Architectural and Functional Comparisons between Prior Works and Proposed FG-DSA*

| Previous Model | Limitation | How FG-DSA Solves It |
|---|---|---|
| FCN for Semantic Segmentation (Long et al., 2015) | Coarse prediction resolution due to spatial downsampling | Recovers high-frequency details via frequency-guided lightweight attention |
| U-Net (Ronneberger et al., 2015) | Loss of border resolution from unpadded convolutions and reliance on custom loss weighting for touching object | Restores fine spatial boundaries and sharpens object transitions via 2D-FFT frequency-guided attention without relying on custom loss weighting |
| FGA (Chaudhuri et al., 2024) | Uses FFT only for global counting, not fine-grained edge recovery | Explicitly extracts high-frequency edges and textures to restore sharp boundary details lost during downsampling |
| FDConv (Chen et al., 2025) | Uses heavy Fourier dynamic convolutions, making it too slow for lightweight networks | Combines frequency-domain priors with lightweight depthwise separable operations to enhance edge representations, with additional parameter and computational costs |
| DSA (Levy, 2026) | Designed to save GPU memory for text generation, but ignores 2D image boundaries and spatial details | Brings sparse attention to vision, boosting both visual understanding (segmentation) and generation (diffusion) |

### (I.3) Novelty of FG-DSA

The innovation in FG-DSA (Figure 1) is introducing a frequency-guided attention mechanism that explicitly couples high-frequency spatial information with lightweight feature interaction, rather than treating frequency enhancement and attention as separate components. FG-DSA uses a parameter-free 2D-FFT high-pass filter to extract high-frequency priors associated with object boundaries and fine structure, then uses these priors to guide depthwise separable attention toward spatial regions where semantic information is most at risk of being lost in fine detail. It provides a way to selectively enhance high-frequency information while maintaining relatively lightweight feature interactions. Beyond discriminative tasks, we also investigate the applicability of FG-DSA in a generative U-Net for single-step image restoration. This study provides an initial assessment of whether frequency-guided attention can improve reconstruction quality and preserve high-frequency details in a generative setting.

**Figure 1**
*Novelty of FG-DSA*

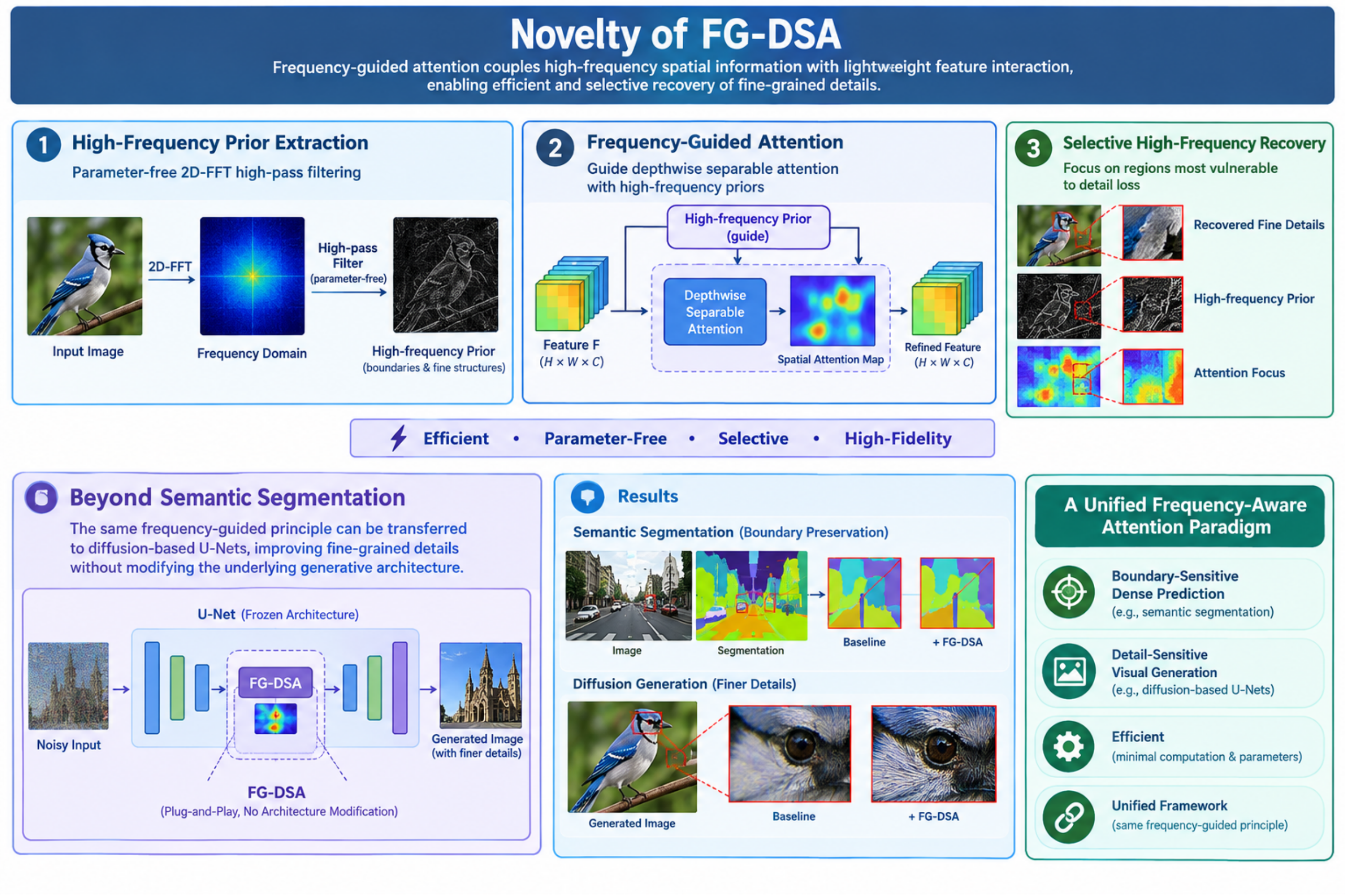

## II. Research Questions

**Research Question 1:** How effectively does the frequency-guided prior in FG-DSA improve dense visual representations and prevent boundary degradation in lightweight segmentation models?

**Research Question 2:** Can FG-DSA be integrated into a generative U-Net for single-step image restoration to improve reconstruction quality and preserve high-frequency image details while maintaining reasonable computational efficiency?

## III. Methodology

To address how frequency-guided priors prevent boundary degradation and refine spatial representations in lightweight dense prediction models, we propose a dual-branch mechanism combining a 2D-FFT High-Pass Filter with a Linear Depthwise Separable Attention module.

### (III.1) 2D-FFT High-Pass Prior Extraction

Repeated spatial downsampling in lightweight backbones causes severe loss of high-frequency details, such as edges, contours, and fine textures. Let $X \in \mathbb{R}^{C \times H' \times W'}$ denote an intermediate spatial feature map. Exploiting the conjugate symmetry of real-valued signals, we apply a 2D Real Fast Fourier Transform (FFT) ($\mathfrak{F}_{rfft2}$) across the spatial dimensions to project $X$ into the frequency domain:

$$\tilde{F}(u, v) = \mathfrak{F}_{rfft2}(X) \in \mathbb{C}^{C \times H' \times (\lfloor \frac{W'}{2} \rfloor + 1)}.$$

- $X$: Input spatial feature map tensor of shape $C \times H' \times W'$.

- $C$: Number of feature channels.
- $H', W'$: Spatial height and width of the intermediate feature map, respectively.
- $\mathfrak{F}_{rfft2}(\cdot)$: 2D Real FFT operator that maps real spatial features into the conjugate-symmetric frequency domain.
- $\tilde{F}(u, v)$: Complex-valued Fourier spectrum at normalized frequency coordinates$(u, v)$, lying in the complex space $\mathbb{C}$ with width $\lfloor \frac{W'}{2} \rfloor + 1$, where $\lfloor \cdot \rfloor$ denotes the floor function.
- $u \in [-1,1], v \in [0,1]$: Normalized vertical and horizontal spatial frequency coordinates, respectively.

To isolate sharp structural contours from low-frequency background semantics, we define a radial binary high-pass filter mask $\mathscr{H}_{HP}(u, v)$ parameterized by normalized cutoff radius $r_{cutoff}$:

$$\mathscr{H}_{HP}(u, v) = \begin{cases} 1, & \text{if} \sqrt{u^2 + v^2} > r_{cutoff}, \\ 0, & \text{otherwise.} \end{cases}$$

Filtering the spectrum via element-wise multiplication ( $\odot$ ) and applying the Inverse 2D Real FFT ($\mathfrak{F}_{irfft2}$) recovers the spatial high-frequency prior $X_{high_freq}$:

$$X_{high_freq} = \mathfrak{F}_{irfft2}\big(\tilde{F}(u, v) \odot \mathscr{H}_{HP}(u, v)\big) \in \mathbb{R}^{C \times H' \times W'}.$$

To adaptively control how much spectral edge information is injected into each feature channel, a Sigmoid-activated $1 \times 1$ convolution, a learnable channel-wise and spatially adaptive gating map $G$, producing the frequency-enhanced feature representation $X_{freq}$ via residual fusion:

$$G = \sigma\big(\text{Conv}_{1\times1}(X_{high_freq})\big), \quad X_{freq} = X + G \odot X_{high_freq}.$$

- $\mathscr{H}_{HP}(u, v)$: Binary radial high-pass mask that preserves frequency components outside the normalized cutoff radius $r_{cutoff}$ and suppresses low-frequency components inside it (set to 0).
- $\odot$: Element-wise multiplication.
- $\mathfrak{F}_{irfft2}(\cdot)$: Inverse 2D Real FFT that reconstructs the real-valued spatial tensor $X_{high_freq}$ from the filtered spectrum.
- $\text{Conv}_{1\times1}(\cdot)$ : Point-wise $1 \times 1$ convolutional learning cross-channel gating weights.
- $\sigma(\cdot)$: Sigmoid activation function, $\sigma(z) = \frac{1}{1 + e^{-z}}$, bounding channel gating weights $G \in (0,1)^{C \times H' \times W'}$.
- $X_{freq}$: Frequency-enhanced feature map obtained via gated residual addition.

**(III.2) Linear Depthwise Separable Attention (DSA)**

Standard self-attention incurs a quadratic $\mathcal{O}((H'W')^2)$ computational complexity with respect to spatial resolution, making it prohibitive for lightweight networks. Instead, FG-DSA projects the frequency-enhanced feature $X_{freq}$ into Query ($Q$), Key ($K$), and Value ($V$) tensors using a unified $1 \times 1$ point-wise convolution:

$$[Q, K, V] = \text{Split}\big(\text{Conv}_{1\times1}(X_{freq})\big), \quad Q, K, V \in \mathbb{R}^{C \times H' \times W'}.$$

To achieve linear complexity $\mathcal{O}(C \cdot H'W')$, $Q$, and $K$ are normalized independently along orthogonal spatial axes:

$$Q_{attn} = \mathrm{Softmax}_{W'}(Q), \quad K_{attn} = \mathrm{Softmax}_{H'}(K).$$

Spatial dimensions are then contracted to aggregate a global channel descriptor $C_c \in \mathbb{R}$ for each channel $c \in \{1,\ldots,C\}$, which modulates $Q_{attn}$ to produce the global attention map $O_{attn}$:

$$C_c = \sum_{h=1}^{H'} \sum_{w=1}^{W'} K_{attn}(c,h,w) \cdot V(c,h,w), \quad O_{attn}(c,h,w) = C_c \cdot Q_{attn}(c,h,w).$$

Finally, a $3 \times 3$ depthwise convolution ($\mathrm{DWConv}_{3\times3}$) followed by a $1 \times 1$ point-wise projection injects localized neighborhood context, yielding the refined bottleneck output via a residual connection:

$$Y = \text{FG-DSA}(X) = X + \mathrm{Conv}_{1\times1}\big(\mathrm{DWConv}_{3\times3}(O_{attn})\big).$$

- $\mathrm{Split}(\cdot)$: Channel-wise partitioning operator dividing the $3C$-channel projected tensor equally into Query ($Q$), Key ($K$), and Value ($V$) tensors.
- $\mathrm{Softmax}_{W'}(\cdot)$, $\mathrm{Softmax}_{H'}(\cdot)$: Softmax normalization applied independently along the spatial width ($W'$) and spatial height ($H'$) dimensions, yielding normalized attention weights $Q_{attn}$ and $K_{attn}$.
- $c \in \{1,\ldots,C\}$: Channel index.
- $h \in \{1,\ldots,H'\}, w \in \{1,\ldots,W'\}$: Spatial row (height) and column (width) indices, respectively.
- $C_c$: Scalar global context descriptor aggregated for channel $c$.
- $O_{attn}$: Intermediate global attention feature map of shape $C \times H' \times W'$.
- $\mathrm{DWConv}_{3\times3}(\cdot)$: $3 \times 3$ Depthwise Convolution applying spatial filtering independently per channel to capture local neighborhood structure.
- $Y$: Final refined output tensor of the FG-DSA module, preserving the input dimensions $C \times H' \times W'$.

**(III.3) Architectural Versatility Across Discriminative and Generative Backbones**

Because FG-DSA preserves tensor dimensions ($Y \in \mathbb{R}^{C\times H'\times W'}$) and requires no task-specific loss modifications, it can be seamlessly inserted into the bottleneck or decoder stages of U-shaped encoder-decoder architectures:

- **Discriminative & Single-Step Generative U-Nets:** FG-DSA is integrated directly at the encoder bottleneck to counteract deep-downsampling blur before decoder upsampling.
- **Extension to Multi-Step Diffusion U-Nets:** In iterative denoising diffusion frameworks, FG-DSA can be placed after skip-connection fusion and generalized with a time-conditioned frequency threshold:

$$h_{l+1} = \text{FG-DSA}\big(\mathrm{UpSample}(h_l) + h_{skip}\big), \quad r_{cutoff}(t) = \sigma(W_t \cdot e(t) + b_t).$$

- $h_l$: Input feature map coming from the preceding lower-resolution decoder stage $l$.
- $h_{l+1}$: Refined output feature map of decoder stage $l+1$.
- $\mathrm{UpSample}(\cdot)$: Spatial upsampling operator matching the spatial resolution of stage $l+1$.
- $h_{skip}$: Skip-connection feature map routed directly from the corresponding encoder stage.

- $t \in \{0,\ldots,T\}$: Current diffusion denoising time-step (ranging from maximum noise step $T$ down to 0).
- $e(t)$: Temporal embedding vector encoding diffusion step $t$.
- $W_t, b_t$: Learnable projection weight matrix and bias parameter mapping $e(t)$ to scalar logit.
- $r_{cutoff}(t) \in (0,1)$: Dynamic normalized cutoff radius conditioned on time-step $t$, enabling the model to suppress high-frequency noise early in denoising ($t \to T$) while sharpening structural edges near $t \to 0$.

## IV. Models

### (IV.1) Baseline U-Net Model Formulation

The baseline U-Net architecture is a symmetric lightweight encoder-decoder with skip connections to preserve multi-scale spatial information.

### (IV.1.1) Encoder Stage

Let $X \in \mathbb{R}^{3\times H\times W}$ denote the input Red, Green, and Blue (RGB) image: $H$ denotes the spatial height of the input image in pixels, and $W$ denotes the spatial width of the input image in pixels. The encoder consists of three hierarchical convolutional stages that progressively downsample spatial resolution by a factor of 2 at each stage:

$$\mathbf{X}_1 = \mathrm{ReLU}\Big(\mathrm{BN}\big(\mathrm{Conv}_{3\times3}(X)\big)\Big), \quad \mathbf{X}_1 \in \mathbb{R}^{32\times\frac{H}{2}\times\frac{W}{2}},$$

$$\mathbf{X}_2 = \mathrm{ReLU}\Big(\mathrm{BN}\big(\mathrm{Conv}_{3\times3}(\mathbf{X}_1)\big)\Big), \quad \mathbf{X}_2 \in \mathbb{R}^{64\times\frac{H}{4}\times\frac{W}{4}},$$

$$\mathbf{X}_3 = \mathrm{ReLU}\Big(\mathrm{BN}\big(\mathrm{Conv}_{3\times3}(\mathbf{X}_2)\big)\Big), \quad \mathbf{X}_3 \in \mathbb{R}^{128\times\frac{H}{8}\times\frac{W}{8}},$$

where $\mathrm{Conv}_{3\times3}$ denotes a standard $3\times3$ convolution with stride $s = 2$ and padding $p = 1$, and BN represents Batch Normalization.

### (IV.1.2) Baseline Bottleneck and Decoder Stage

In the baseline U-Net, the bottleneck feature map $\mathbf{X}_3$ is directly routed to the decoder without frequency-domain enhancement. The decoder progressively restores spatial resolution via transposed convolutions while aggregating shallow feature maps through additive skip connections:

$$\mathbf{D}_2 = \mathrm{ReLU}\Big(\mathrm{BN}\big(\mathrm{ConvTranspose}_{4\times4}(\mathbf{X}_3)\big)\Big) + \mathbf{X}_2, \quad \mathbf{D}_2 \in \mathbb{R}^{64\times\frac{H}{4}\times\frac{W}{4}},$$

$$\mathbf{D}_1 = \mathrm{ReLU}\Big(\mathrm{BN}\big(\mathrm{ConvTranspose}_{4\times4}(\mathbf{D}_2)\big)\Big) + \mathbf{X}_1, \quad \mathbf{D}_1 \in \mathbb{R}^{32\times\frac{H}{2}\times\frac{W}{2}},$$

where $\mathrm{ConvTranspose}_{4\times4}$ denotes a transposed convolution with kernel size 4, stride 2, and padding 1.

### (IV.1.3) Segmentation Head

The final prediction logits $\mathbf{Y}_{\mathrm{base}} \in \mathbb{R}^{K\times H\times W}$ for $K$ semantic categories are generated by:

$$\mathbf{Y}_{\mathrm{base}} = \mathrm{Conv}_{1\times1}\Big(\mathrm{ConvTranspose}_{4\times4}(\mathbf{D}_1)\Big).$$

### (IV.2) U-Net with Proposed Frequency-Guided Depthwise Separable Attention (FG-DSA)

To prevent boundary degradation caused by repeated downsampling, we integrate the proposed FG-DSA module (formulated in Section III) directly into the encoder bottleneck ($C = 128$, $H' = \dfrac{H}{8}$, and $W' = \dfrac{W}{8}$):

$$\mathbf{X}_3^{\mathrm{FG-DSA}} = \mathscr{F}_{\mathrm{FG-DSA}}(\mathbf{X}_3).$$

#### (IV.2.1) 2D Real FFT High-Pass Prior

For the 2D real fast Fourier Transform (2D real FFT), the spatial bottleneck representation $\mathbf{X}_3 \in \mathbb{R}^{C\times H'\times W'}$, where channel dimensions $C = 128$, $H' = \dfrac{H}{8}$, and $W' = \dfrac{W}{8}$ are transformed into the conjugate-symmetric complex frequency domain. We can represent the 2D real fast Fourier Transform operator as:

$$\tilde{\mathbf{F}}(u, v) = \mathscr{F}_{\mathrm{rfft2}}(\mathbf{X}_3), \quad \tilde{\mathbf{F}}(u, v) \in \mathbb{C}^{C\times H'\times(\lfloor \frac{W'}{2} \rfloor+1)},$$

where $\mathscr{F}_{\mathrm{rfft2}}(\,\cdot\,)$ denotes the 2D Real FFT operator, $\mathbb{C}$ represents the complex number space, and $\lfloor\,\cdot\,\rfloor$ is the floor function.

Let $u \in [-1,1]$ and $v \in [0,1]$ represent normalized spatial frequency coordinates. The spectral radius from the frequency center is given by

$$R(u, v) = \sqrt{u^2 + v^2}.$$

Given a predefined cutoff threshold $r_{\mathrm{cutoff}}$, the binary high-pass (HP) mask $\mathscr{M}_{\mathrm{HP}}(u, v)$ is defined as:

$$\mathscr{M}_{\mathrm{HP}}(u, v) = \begin{cases} 1, & \text{if } \sqrt{u^2 + v^2} > r_{\mathrm{cutoff}}, \\ 0, & \text{otherwise} \end{cases}$$

where a mask value of 1 retains the high-frequency structural edge components beyond the cutoff radius, and a mask value of 0 suppresses the low-frequency background harmonics.

Applying the high-pass mask suppresses low-frequency background information and retains sharp edge components. The spatial high-frequency representation is recovered via Inverse 2D Real FFT ($\mathscr{F}_{\mathrm{irfft2}}$):

$$\tilde{\mathbf{F}}_{\mathrm{HP}}(u, v) = \tilde{\mathbf{F}}(u, v) \odot \mathscr{M}_{\mathrm{HP}}(u, v),$$

$$\mathbf{X}_{\mathrm{high_freq}} = \mathscr{F}_{\mathrm{irfft2}}(\tilde{\mathbf{F}}_{\mathrm{HP}}(u, v)) \in \mathbb{R}^{C\times H'\times W'},$$

where $\mathscr{F}_{\mathrm{irfft2}}$ is an inverse 2D Real FFT, and $\mathbf{X}_{\mathrm{high_freq}}$ is the reconstructed spatial high-frequency representation.

Let $\mathrm{Conv}_{1\times1}(\,\cdot\,)$ be a $1 \times 1$ convolution operation that learns channel-wise modulation weights from the high-frequency representation to dynamically modulate high-frequency contributions. A Sigmoid gating mechanism computes channel-wise modulation weights for residual fusion $\mathbf{G}$:

$$\mathbf{G} = \sigma\big(\mathrm{Conv}_{1\times1}(\mathbf{X}_{\mathrm{high_freq}})\big),$$

$$\mathbf{X}_{\mathrm{freq}} = \mathbf{X}_3 + \mathbf{G} \odot \mathbf{X}_{\mathrm{high_freq}},$$

where $\sigma(\,\cdot\,)$ denotes the Sigmoid activation function, and $\odot$ represents the element-wise product.

### (IV.2.2) Depthwise Separable Linear Attention

Queries $\mathbf{Q}$, Keys $\mathbf{K}$, and Values $\mathbf{V}$ are generated using a $1 \times 1$ point-wise convolution:

$$[\mathbf{Q}, \mathbf{K}, \mathbf{V}] = \mathrm{Split}\Big(\mathrm{Conv}_{1\times1}(\mathbf{X}_{\mathrm{freq}})\Big), \quad \mathbf{Q}, \mathbf{K}, \mathbf{V} \in \mathbb{R}^{C\times H'\times W'}.$$

To avoid the quadratic $\mathcal{O}((H'W')^2)$ complexity of standard self-attention, $\mathbf{Q}$ and $\mathbf{K}$ are normalized along orthogonal spatial dimensions:

$$\mathbf{Q}_{\mathrm{attn}} = \mathrm{Softmax}_{W'}(\mathbf{Q}), \quad \mathbf{K}_{\mathrm{attn}} = \mathrm{Softmax}_{H'}(\mathbf{K}).$$

Spatial dimensions are contracted to construct a global channel descriptor $\mathbf{C}_c \in \mathbb{R}^1$ for each channel $c$, reducing the attention complexity to a linear scale $\mathcal{O}(C \cdot H'W')$ of the proposed attention mechanism with respect to the spatial area $(H'W')$, by passing the quadratic complexity $\mathcal{O}((H'W')^2)$ of standard self-attention:

$$\mathbf{C}_c = \sum_{h=1}^{H'} \sum_{w=1}^{W'} \mathbf{K}_{\mathrm{attn}}(c, h, w) \cdot \mathbf{V}(c, h, w),$$

$$\mathbf{O}_{\mathrm{attn}}(c, h, w) = \mathbf{C}_c \cdot \mathbf{Q}_{\mathrm{attn}}(c, h, w).$$

A $3 \times 3$ depthwise convolution followed by a $1 \times 1$ point-wise projection injects localized spatial context into the attention map:

$$\mathbf{X}_3^{\mathrm{FG-DSA}} = \mathbf{X}_3 + \mathrm{Conv}_{1\times1}\Big(\mathrm{DWConv}_{3\times3}(\mathbf{O}_{\mathrm{attn}})\Big).$$

- $\mathbf{X}_3^{\mathrm{FG-DSA}}$ : The final output feature map tensor of the bottleneck layer after being enhanced by the proposed FG-DSA module ($\mathbf{X}_3^{\mathrm{FG-DSA}} \in \mathbb{R}^{C\times H'\times W'}$).

- $\mathbf{X}_3$: The original input bottleneck feature map tensor fed into the FG-DSA module from the encoder stage ($\mathbf{X}_3 \in \mathbb{R}^{C\times H'\times W'}$).

- $\mathbf{O}_{\mathrm{attn}}$: The intermediate output feature tensor generated by the Depthwise Separable Linear Attention layer in the previous step ($\mathbf{O}_{\mathrm{attn}} \in \mathbb{R}^{C\times H'\times W'}$).

- $\mathrm{DWConv}_{3\times3}$: A $3 \times 3$ Depthwise Convolution operator that applies spatial filtering independently to each feature channel, injecting localized spatial context with low parameter overhead.

- $\mathrm{Conv}_{1\times1}$: A $1 \times 1$ Point-wise Convolution operator used to perform cross-channel linear projection and feature mixing.

### (IV.2.3) Proposed Decoder and Loss Formulation

The enhanced feature representation $\mathbf{X}_3^{\mathrm{FG-DSA}}$ is passed to the proposed decoder stage:

$$\mathbf{D}_2^{\mathrm{prop}} = \mathrm{ReLU}\Big(\mathrm{BN}\big(\mathrm{ConvTranspose}_{4\times4}(\mathbf{X}_3^{\mathrm{FG-DSA}})\big)\Big) + \mathbf{X}_2,$$

$$\mathbf{Y}_{\mathrm{proposed}} = \mathrm{Conv}_{1\times1}\Big(\mathrm{ConvTranspose}_{4\times4}\big(\mathrm{ReLU}(\mathrm{BN}(\mathrm{ConvTranspose}_{4\times4}(\mathbf{D}_2^{\mathrm{prop}}))) + \mathbf{X}_1\big)\Big).$$

- $\mathbf{X}_3^{\mathrm{FG-DSA}}$: Enhanced bottleneck feature representation produced by the proposed FG-DSA module.

- $\mathbf{D}_2^{\text{prop}}$: Intermediate decoder feature representation obtained by upsampling the enhanced bottleneck features and fusing them with the encoder features $\mathbf{X}_2$.

- $\mathbf{Y}_{\text{proposed}}$: Final predicted output generated by the proposed decoder.

- $\mathbf{X}_2$: Encoder feature representation from the second stage, used for skip-connection fusion with the decoder features at the corresponding spatial resolution.

- $\mathbf{X}_1$: Encoder feature representation from the first stage, used for skip-connection fusion to recover fine-grained spatial details.

- $\text{ConvTranspose}_{4\times4}(\cdot)$: A $4 \times 4$ transposed convolution operation, typically used to upsample feature maps and increase their spatial resolution.

- $\text{BN}(\cdot)$: Batch Normalization, which normalizes feature activations to improve training stability and convergence.

- $\text{ReLU}(\cdot)$: Rectified Linear Unit activation function, defined as $\text{ReLU}(x) = \max(0,x)$, which introduces nonlinearity by setting negative values to zero.

- $\text{Conv}_{1\times1}(\cdot)$: A $1 \times 1$ convolution operation used to map the decoder features to the desired number of output channels, such as the number of segmentation classes.

The network is optimized end-to-end using the standard multi-class cross-entropy loss $\mathscr{L}_{\text{CE}}(\mathbf{Y}, \mathbf{T})$ that measures the discrepancy between the predicted class probability and the ground-truth labels, computed over all spatial pixels $N = H \times W$:

$$\mathscr{L}_{\text{CE}}(\mathbf{Y}, \mathbf{T}) = -\frac{1}{N}\sum_{i=1}^{N}\sum_{k=0}^{K-1}\mathbf{T}_{i,k}\log\left(\frac{\exp(\mathbf{Y}_{i,k})}{\sum_{j=0}^{K-1}\exp(\mathbf{Y}_{i,j})}\right),$$

where $\mathbf{T}_{i,k} \in \{0,1\}$ represents the ground-truth one-hot target indicator for class $k$ at pixel index $i$, and $\mathbf{Y}_{i,k}$ is the predicted logit for class $k$ at pixel $i$, representing the raw score before Softmax normalization.

## V. Datasets

### (V.1) Dataset 1 - CamVid Semantic Segmentation Benchmark

To assess discriminative dense prediction in Research Question 1, we use the Cambridge-Driving Labeled Video Database (CamVid) (stonewst, 2026) as our first benchmark dataset. The data was captured from the driver's point of view in a variety of complex urban street scenes and comprises high-resolution RGB images, resized to $480 \times 360$, and dense pixel-level semantic annotation maps. As per the standard 11-class split, with large homogeneous background classes such as sky, building, and road, and a number of thin, high-frequency foreground classes such as pole, fence, pedestrian, bicyclist, and sign symbol, the dataset is split into 367 training images, 101 validation images, and 233 test images, with model training and quantitative evaluation conducted on the standard training and validation splits, respectively. The labeled benchmark provides an unbiased evaluation of FG-DSA's ability to prevent boundary feature degradation along the contours of slender objects.

### (V.2) Dataset 2- Self-Constructed High-Frequency Image Restoration Dataset

To properly assess cross-task generalization to generative architectures in Research Question 2 without introducing any domain shift, we create a second, task-specific dataset for high-frequency image restoration and super-resolution reconstruction. It is built on the unlabeled urban street imagery, including 367 training pairs and 101 validation pairs, all resized to 480 × 360 pixels, and we apply a scripted degradation pipeline to create the paired low-quality inputs and clean high-resolution targets. Each clean target is first downsampled by a factor of 2 via area interpolation, then passed through a 5 × 5 Gaussian blur and upscaled back to the original resolution via linear interpolation to simulate severe degradation of high-frequency spatial content. The resulting self-constructed paired dataset is used to directly test FG-DSA's modular capability to recover sharp structural textures and fine perceptual details of the original image during generative upsampling.

## VI. Result (Experiments and Analysis)

### (VI.1) Analysis of the Hyper-Parameter Settings

To ensure reproducibility and computational efficiency, all experiments are run with fixed random seeds (seed = 75 for model initialization and seed = 42 for data shuffling) under deterministic cuDNN execution. The input images are resized to 480 × 360 and downsampled 8 × across 3 encoder stages to produce a 60x45 bottleneck feature map, which balances global context with low memory cost. Within FG-DSA, the 2D-FFT high-pass filter uses a normalized cutoff radius of $r_{cutoff} = 0.15$ to suppress low-frequency background illumination while retaining fine object contours, followed by a 3 × 3 depthwise convolution for local refinement. Both the baseline and proposed models are trained for 15 epochs using the AdamW optimizer (learning rate = $1 \times 10^{-3}$, weight decay = $1 \times 10^{-4}$) with Automatic Mixed Precision (FFT operations cast to FP32 for stability) to avoid overfitting on the 367-sample training set. Task-specific configurations use a batch size of 32 for Research Question 1 segmentation, evaluated in the 3-pixel and 5-pixel morphological edge bands, and a batch size of 16 for Research Question 2 generative restoration, using a composite objective to balance pixel fidelity with high-frequency edge sharpness.

### (VI.2) Baseline U-Net & U-Net with FG-DSA

**Baseline U-Net and Segmentation U-Net with FG-DSA:** For the semantic segmentation task in answer to Research Question 1, we use the classic U-Net as our baseline (Baseline U-Net) and incorporate our FG-DSA module into the U-Net bottleneck to create the Segmentation U-Net with FG-DSA.

**Baseline U-Net and Generative U-Net with FG-DSA:** For the generative reconstruction task in answer to Research Question 2, we use the classic U-Net as our baseline (Baseline U-Net) and incorporate our FG-DSA module into its bottleneck to create the Generative U-Net with FG-DSA.

### (VI.3) Result of Research Question 1

We assess the ability of the frequency-guided prior within the FG-DSA module to improve the density of the feature representations and to combat edge degradation in the model architecture. The epoch-by-epoch training loss comparison (Table 2) & (Figure 2) is shown over 15 epochs. The empirical dynamics demonstrate a clear two-phase optimization progression that validates the design of FG-DSA:

(1) Gating Parameter Warm-Up Phase (Epochs 0-4): Over the first four epochs, the baseline U-Net actually has a slightly lower training loss than the proposed U-Net with FG-DSA (1.8152 vs 1.9108 at epoch 1, 0.9233 vs 0.9383 at Epoch 04), giving positive loss differences of +0.0956 and +0.0150, respectively. This is, of course, the expected parameter warm-up period, during which the Sigmoid frequency gating learns to filter out low-frequency noise and pick out high-frequency boundary components.

(2) Crossover and Superior Dense Representation (Epochs 5-15): At Epoch 05, we have a major crossover with the proposed U-Net with FG-DSA model dropping to a loss of 0.8322, which is below the baseline U-Net's 0.8741 (negative loss difference of −0.0419). After that point, the baseline U-Net enters an optimization plateau, with its loss dropping by only 0.2649 over the next ten epochs due to the model's capacity and the spatial smoothing of the convolutions. The proposed U-Net with FG-DSA, on the other hand, continues to optimize at a steady rate, its loss dropping by 0.3735 over the same period.

By Epoch 15, the proposed U-Net with FG-DSA achieves an overall loss floor of 0.4587, outperforming the baseline U-Net (0.6092) with a substantial net loss reduction of 0.1505. This clearly demonstrates that explicitly incorporating Fourier high-pass priors effectively resolves fine-scale spatial semantics and reduces boundary blur, thereby overcoming the representation bottlenecks inherent in lightweight segmentation networks.

**Table 2**
*Epoch-by-Epoch Training Loss Comparison*

| Epoch | Baseline U-Net | U-Net with FG-DSA | Loss Difference |
|---|---|---|---|
| 1 | 1.8152 | 1.9108 | Increase 0.0956 |
| 2 | 1.2129 | 1.3234 | Increase 0.1105 |
| 3 | 1.0208 | 1.0945 | Increase 0.0737 |
| 4 | 0.9233 | 0.9383 | Increase 0.0150 |
| 5 | 0.8741 | 0.8322 | Decrease 0.0419 |
| 6 | 0.8451 | 0.7672 | Decrease 0.0779 |
| 7 | 0.8112 | 0.7132 | Decrease 0.0980 |
| 8 | 0.7829 | 0.6834 | Decrease 0.0995 |
| 9 | 0.7514 | 0.6464 | Decrease 0.1050 |
| 10 | 0.7235 | 0.6056 | Decrease 0.1179 |
| 11 | 0.7164 | 0.5843 | Decrease 0.1321 |
| 12 | 0.6746 | 0.5502 | Decrease 0.1244 |
| 13 | 0.6450 | 0.5210 | Decrease 0.1240 |
| 14 | 0.6385 | 0.4894 | Decrease 0.1491 |
| 15 | 0.6092 | 0.4587 | Decrease 0.1505 |

*Note.* Ablation Study.

**Figure 2**
*Epoch-by-Epoch Training Loss Dynamics*

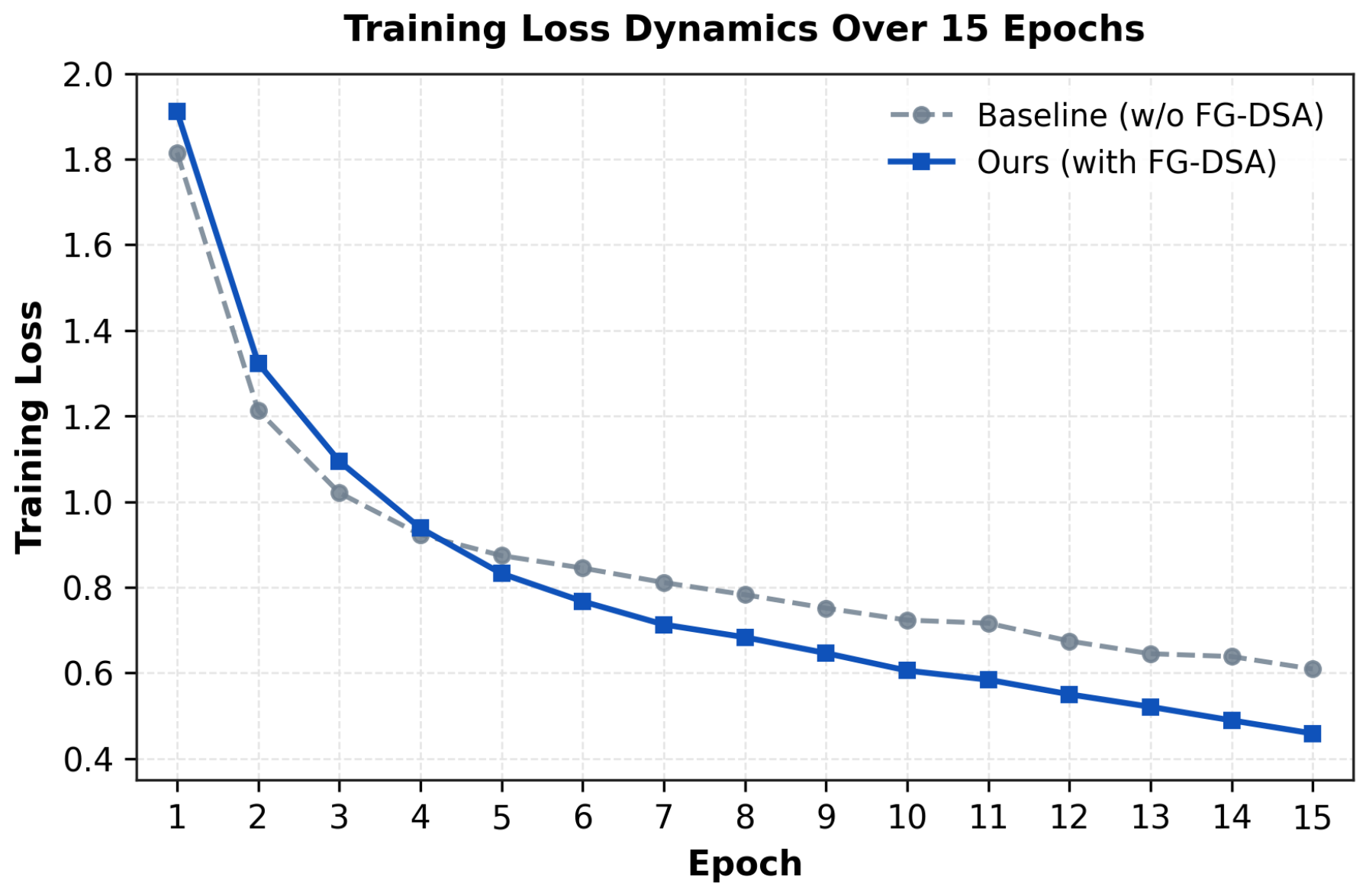

*Note.* Ablation Study.

Moreover, the visual predictions (Figure 3) strongly reinforce these quantitative findings and provide intuitive evidence of enhanced spatial coherence. As shown in the second column, the baseline U-Net is marred by severe spatial fragmentation and noisy misclassifications within the homogeneous areas, such as misclassifying road pixels as building pixels. The U-Net with FG-DSA, on the other hand, produces clean, continuous semantic regions that closely match the ground truth. Crucially, the high-pass Fourier prior preserves thin linear edge features, such as continuous curb lines and lane markings along the road, which are severed or blurred in the baseline U-Net output. The visual comparisons clearly demonstrate that the frequency-guided attention mechanism effectively suppresses low-frequency noise and boundary degradation, allowing lightweight backbones to maintain sharp structural contours.

**Figure 3**

*Visual comparison between the baseline U-Net and U-Net with FG-DSA against Ground Truth*

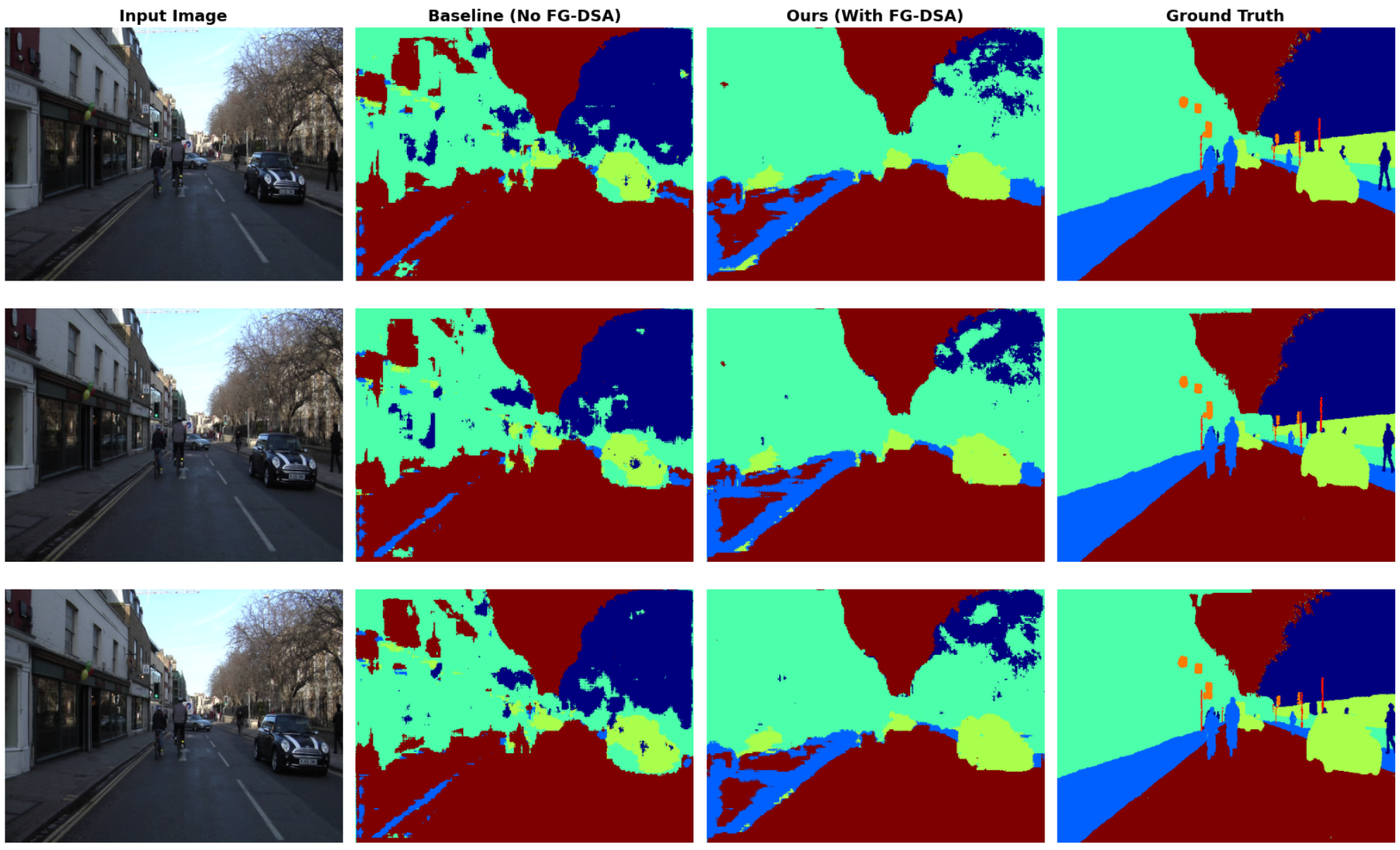

*Note.* Ablation Study.

To properly assess the impact of our FG-DSA, we look at both global semantic accuracy and fine-grained spatial boundary integrity. Standard evaluation metrics computed across all semantic classes are prone to denominator inflation due to large, easily segmented background classes, such as sky and flat road. These large homogeneous regions have very little structural ambiguity, so gains in accuracy along complex object boundaries are effectively diluted.

To reduce this background interference and isolate the high-frequency spatial improvements, we establish a targeted boundary evaluation protocol that focuses on fine-grained, high-frequency structural classes. We also increase the morphological edge band width to ±5px to capture the full spatial transition zone, which is susceptible to feature attenuation due to multi-scale downsampling.

As per the pixel-level and boundary-level performance comparison (Table 3) & (Figure 4), the inclusion of FG-DSA results in consistent improvements in global pixel accuracy across all the samples. Sample 1 goes from 76.6204% to 77.5961% (1,686 pixels fixed), Sample 2 from 77.0828% to 78.8397% (3,036 pixels fixed), and Sample 3 from 77.00% to 79.0648% (3,568 pixels fixed). The overall boundary error reduction across all classes is very modest (0.9985% to 2.4635%) because the denominator is inflated by the smoothness of the background borders. The Hard-Class Boundary Error Reduction Rate (±5px), however, improves substantially, achieving 11.6422% for Sample 1, 11.8777% for Sample 2, and 12.0574% for Sample 3. It shows that the high-frequency spectral prior of FG-DSA is particularly effective at correcting severe degradation along thin, fine object boundaries.

**Table 3**

*Pixel-Level and Boundary-Level Performance Comparison*

| Sample | Pixel Accuracy (%) | | Total Error Pixels | | All-Class Boundary Error (±5px) | | Hard-Class Boundary Error (±5px) | |
|---|---|---|---|---|---|---|---|---|
| | Baseline U-Net | U-Net with FG-DSA | Baseline U-Net | U-Net with FG-DSA | Baseline U-Net | U-Net with FG-DSA | Baseline U-Net | U-Net with FG-DSA |
| 1 | 76.6204 | 77.5961 | 40,400 | 38,714 | 8,212 | 8,130 | 5,858 | 5,176 |
| 2 | 77.0828 | 78.8397 | 39,601 | 36,565 | 8,244 | 8,074 | 5,986 | 5,275 |
| 3 | 77.0000 | 79.0648 | 39,744 | 36,176 | 8,362 | 8,156 | 6,129 | 5,390 |

*Note.* Ablation Study.

**Figure 4**

*Comparative Evaluation of FG-DSA on Individual Samples & Hard Boundaries*

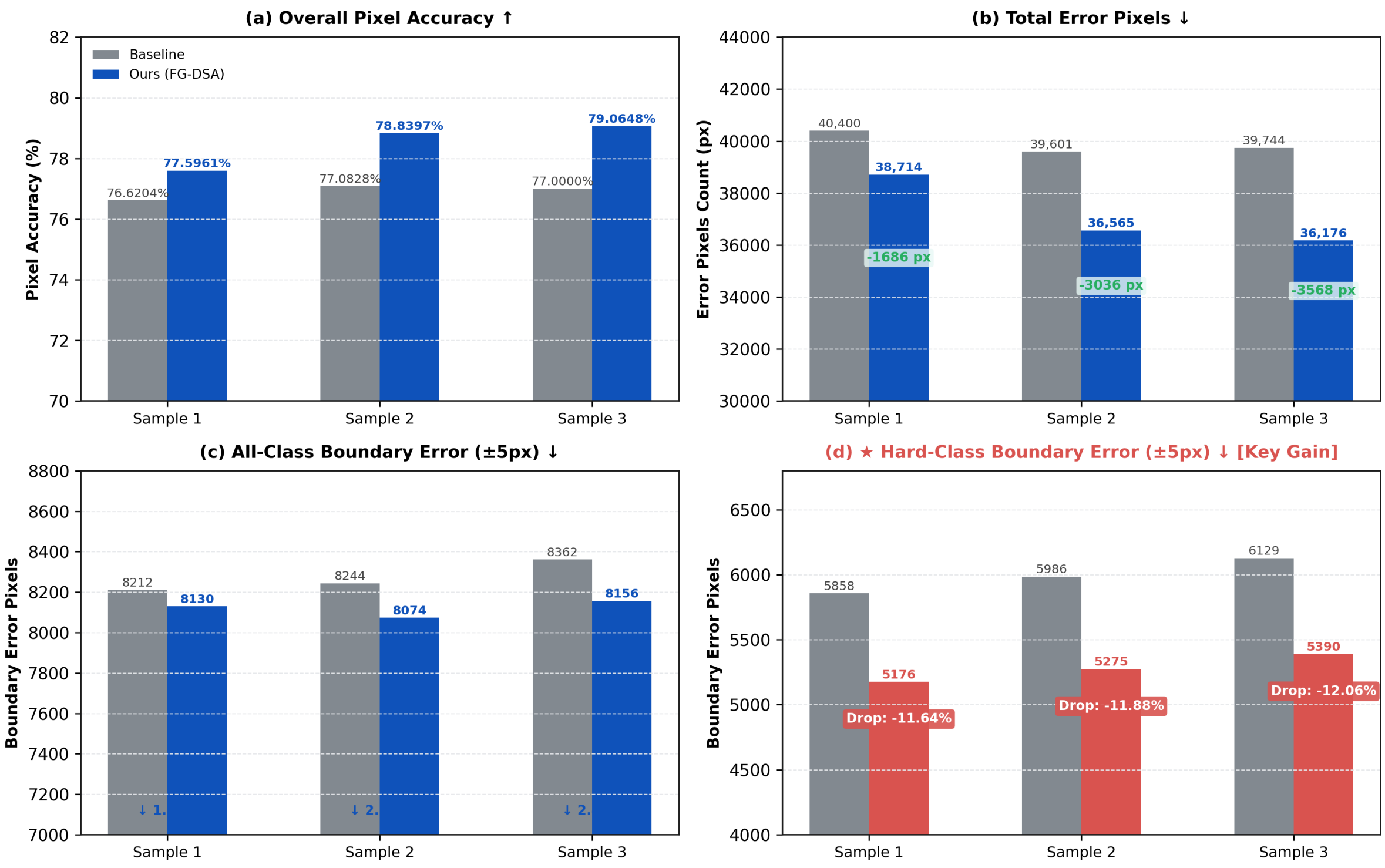

*Note.* Ablation Study.

The pixel-level error map analysis further supports these quantitative results. As shown in the visual comparison (Figure 5). The baseline predictions (column 2, red error regions) exhibit substantial error clustering along dense building silhouettes, tree contours, distant road margins, and fine-grained street features. The error reduction maps (column 4) have the corrected pixels highlighted in green. The spatial distribution of the green pixels shows that FG-DSA can correct large clusters of misclassified pixels along structural boundaries, such as building boundaries, sidewalk edges, and the contours of slender foreground objects.

**Figure 5**

*Error Map Analysis and Spatial Error Reduction*

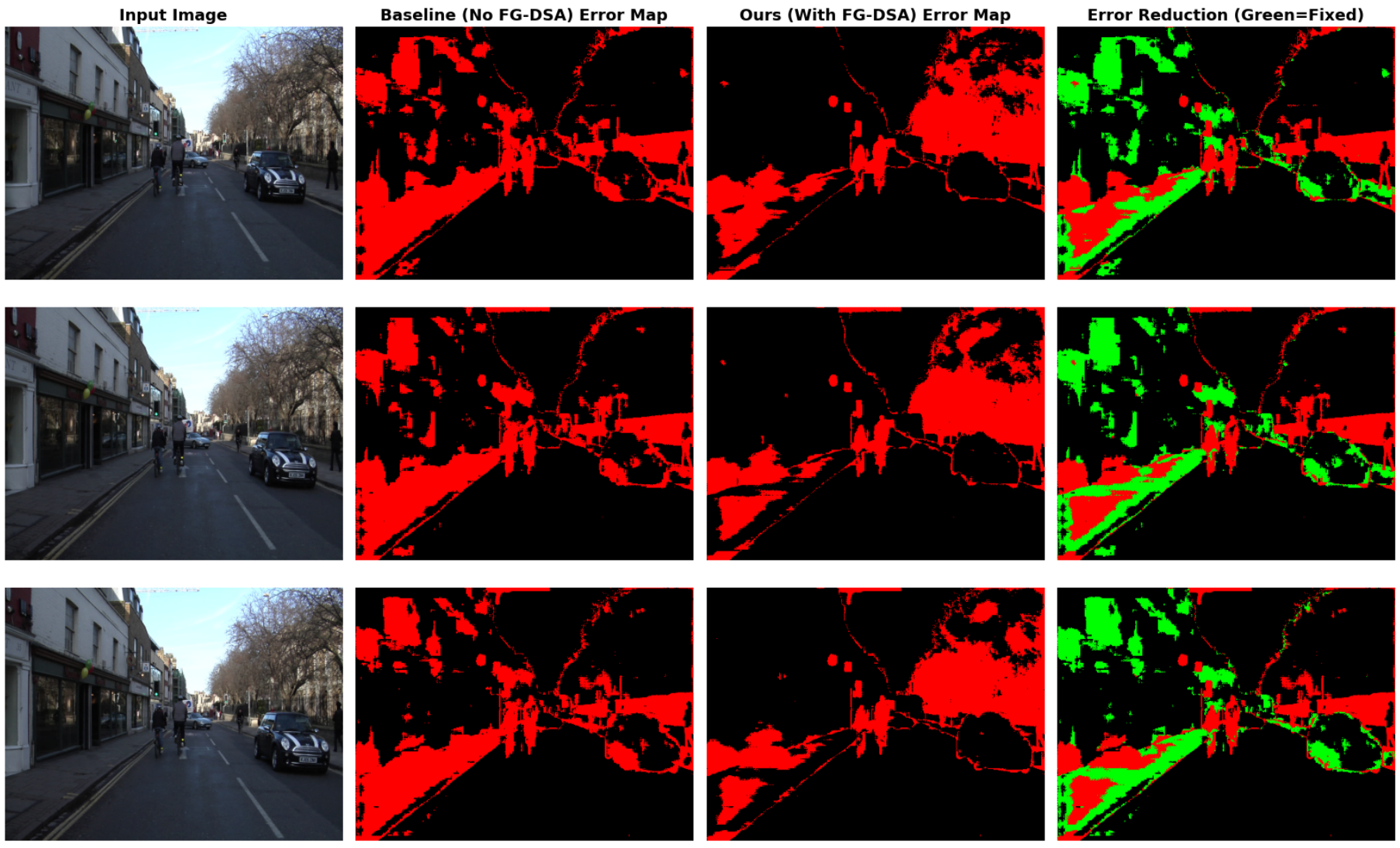

*Note.* Ablation Study.

To investigate the inner mechanism by which FG-DSA mitigates boundary degradation, we examine the bottleneck feature representations before and after integrating the module. Lightweight encoders generally suffer from significant feature attenuation at deeper stages; as a result, activation maps become overly smoothed along the thin, fine-grained contours of objects.

As detailed in the quantitative activation metrics (Table 4), evaluating mean feature intensities exclusively along hard-class boundary zones (±5px band) reveals a substantial increase in feature response. The baseline bottleneck has very low mean boundary responses, from 0.3082 to 0.3717. With FG-DSA, the mean hard class boundary activation increases to 0.4099 (an increase of 10.2794%) in Sample 1, 0.3734 (an increase of 21.1611%) in Sample 2, and 0.4331 (an increase of 22.9595%) in Sample 3. This is a massive gain and clearly demonstrates that FG-DSA dynamically injects high-frequency spectral priors into the deep layers where local detail has been lost.

**Table 4**

*Feature Heatmap & High-Frequency Activation Analysis*

| Sample | Baseline U-Net | U-Net with FG-DSA | Improvement |
|---|---|---|---|
| 1 | 0.3717 | 0.4099 | Increases 10.2794% |
| 2 | 0.3082 | 0.3734 | Increases 21.1611% |
| 3 | 0.3523 | 0.4331 | Increases 22.9595% |

*Note.* Ablation Study.

The feature maps (Figure 6) align with the quantitative improvements. The baseline model (column 2) is extremely over-smoothed, with a very diffuse activation pattern dominated by low-frequency background semantics and high-frequency boundaries (building edges, street lamps, fences, etc.) completely blurred out. The FG-DSA-integrated model (column 3), on the other hand, produces sharp boundary localization with high-contrast activation along fine-grained features before decoder upsampling. By selectively boosting high-frequency spectral components at the bottleneck, we increase hard class boundary feature response by up to 22.96%, effectively combating boundary degradation in lightweight segmentation networks.

**Figure 6**

*Visualization of High-Frequency Feature Enhancement by FG-DSA*

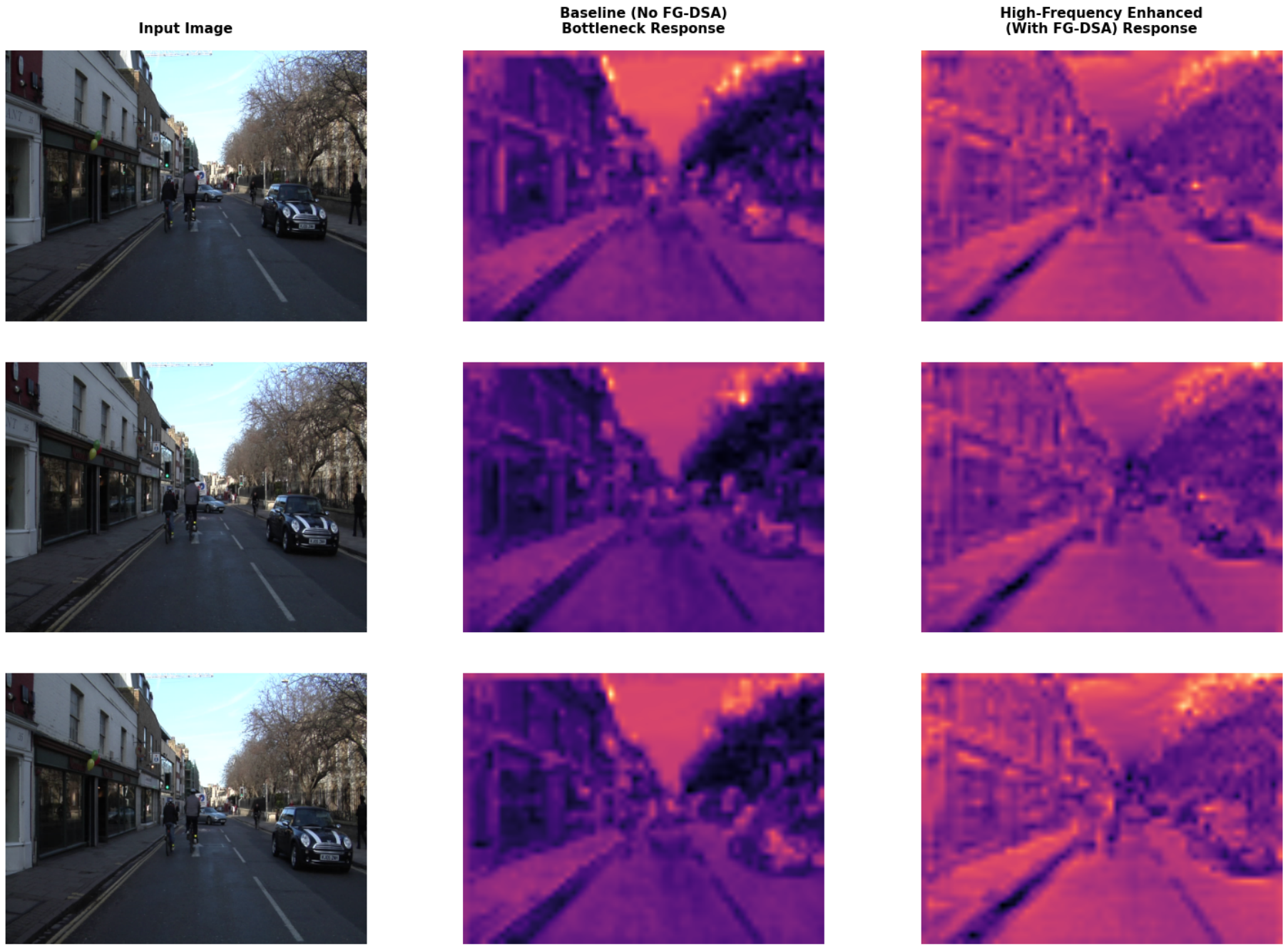

*Note.* Ablation Study.

Our evaluation (Table 5) shows that the frequency-guided prior in FG-DSA effectively boosts dense visual representations and prevents boundary degradation in lightweight segmentation architectures. Including FG-DSA yields a fairly good overall gain of +2.4200 percentage points in Global mIoU (from 37.3673% to 39.7873%). However, across the morphological ±5px boundary band, the gain is modest at +1.1964 percentage points (from 19.5847% to 20.7811%), as the denominator is heavily inflated by large background regions that are very easy to segment. The module's real impact is clear when you remove the background and evaluate only the fine-grained, high-frequency structural classes. In that case, the Hard-Class Boundary mIoU increases by 10.1422 percentage points, from 18.4731% to 28.6153%. This remarkable performance leap shows that FG-DSA counters feature attenuation in lightweight encoders by injecting targeted high-frequency spectral cues into deep bottleneck representations, directly preventing boundary over-smoothing and preserving sharp contours along complex object perimeters.

**Table 5**

*Impact of FG-DSA on Global and Boundary Segmentation Accuracy*

| Metric | Baseline U-Net | U-Net with FG-DSA | Improvement |
|---|---|---|---|
| Global mIoU (%) | 37.3673 | 39.7873 | Increases 2.4200 percentage points |
| All-Class Boundary mIoU (±5px) (%) | 19.5847 | 20.7811 | Increases 1.1964 percentage points |
| Hard-Class Boundary mIoU (±5px) (%) | 18.4731 | 28.6153 | Increases 10.1422 percentage points |

*Note.* Ablation Study.

The segmentation predictions directly validate the quantitative metrics (Figure 7). As clearly demonstrated in the baseline predictions (column 2), the U-Net without FG-DSA suffers from severe fragmented misclassifications and structural over-smoothing along building silhouettes, tree-sky boundaries, and distant street regions. In contrast, the inclusion of FG-DSA (column 3) successfully suppresses the noisy background to produce well-contiguous semantic regions, whilst maintaining the sharp, well-defined contours along the fine-grained structures, such as street lamps, pedestrians, and road boundaries, which are in very close agreement with the ground truth (column 4). It demonstrates the effectiveness of the frequency-guided prior in correcting boundary degradation in the lightweight network.

**Figure 7**
*Visual Comparison of Segmentation Maps*

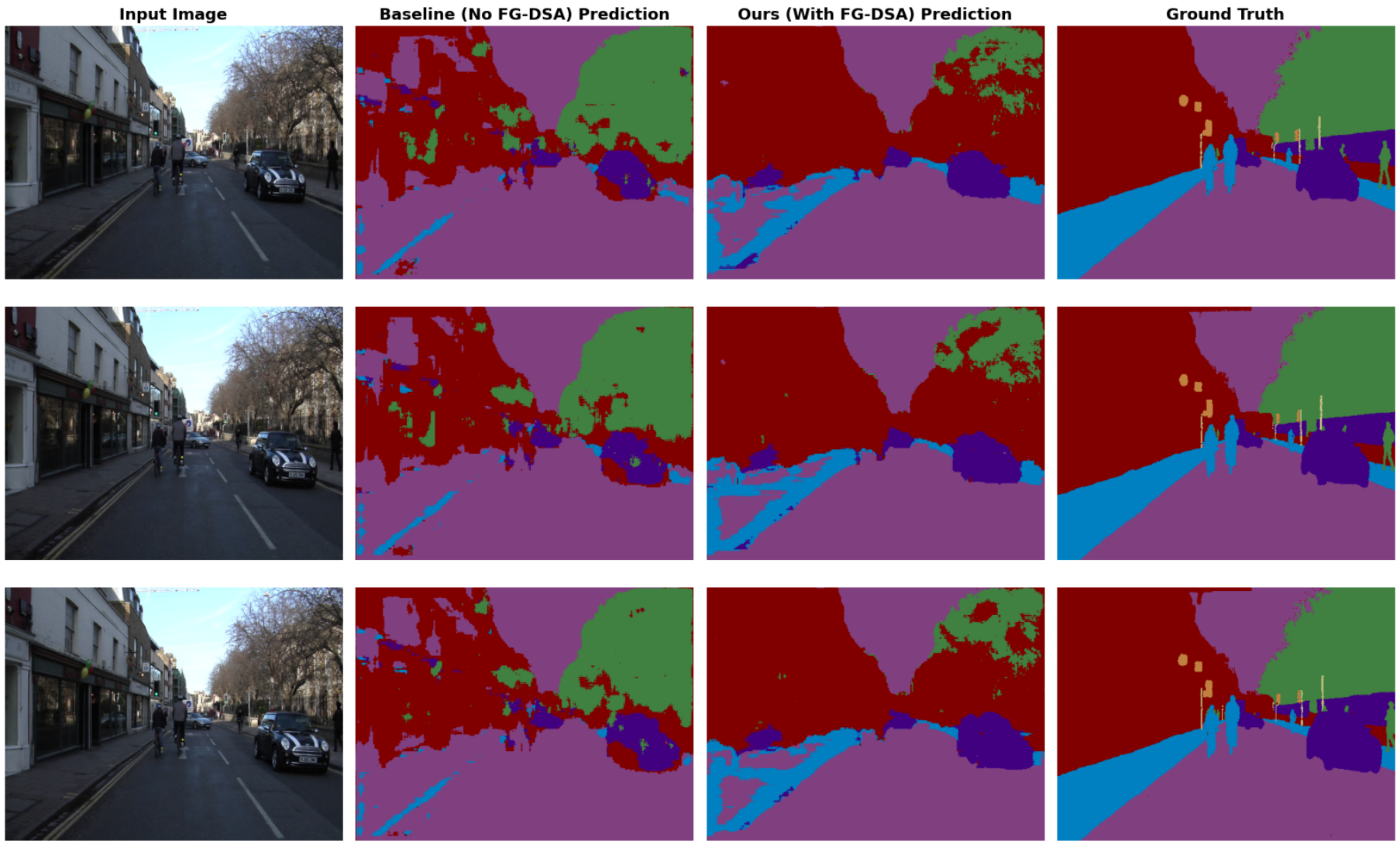

*Note.* Ablation Study.

### (VI.4) Result of Research Question 2

We investigate whether the proposed FG-DSA can generalize to generative architectures to improve the level of spatial detail synthesized while still achieving an optimal accuracy vs. efficiency trade-off. The dataset was pre-loaded into system memory, and we had a Generative U-Net as our baseline architecture. We constructed the proposed variant by inserting FG-DSA into the U-Net bottleneck. Both networks were trained end-to-end for 15 epochs using exactly the same hyperparameters and optimization schedules. In the controlled benchmark, we evaluate the core bottleneck instantiation of FG-DSA with a fixed cutoff radius ($r_{cutoff} = 0.15$) on a single-step generative image restoration task, isolating the spatial-frequency reconstruction gain without the complications of multi-step diffusion sampling schedules. Extending this to time-conditioned $r_{cutoff}(t)$ within full iterative diffusion models is of course a very natural next step for future deployment.

During training, both the U-Net baseline and the U-Net with FG-DSA showed smooth and stable convergence over the 15-epoch period. The epoch-by-epoch training curve and performance comparison (Table 6) & (Figure 8) show that the baseline U-Net starts at Epoch 1 with a training loss of 0.2457, a validation PSNR of 17.9127 dB, and a high-frequency MSE of 0.1668. It finishes at Epoch 15 with a training loss of 0.1156, a PSNR of 24.1974 dB, and a high-frequency MSE of 0.0811. The proposed U-Net with FG-DSA starts at Epoch 1 with a training loss of 0.2595, a PSNR of 17.4538 dB, and a high-frequency MSE of 0.1686. It then rapidly improves in performance, reaching a training loss of 0.1175, a high-frequency MSE of 0.0737 at Epoch 14, and a PSNR of 24.5342dB with a high-frequency MSE of 0.0793 at Epoch 15.

**Table 6**

*Epoch-by-Epoch Convergence Trajectory and Reconstruction Performance Comparison*

| Epoch | Baseline Train Loss | Baseline U-Net Validation PSNR (dB) | Baseline U-Net High-Frequency MSE | U-Net with FG-DSA Train Loss | U-Net with FG-DSA Validation PSNR (dB) | U-Net with FG-DSA High-Frequency MSE |
|---|---|---|---|---|---|---|
| 1 | 0.2457 | 17.9127 | 0.1668 | 0.2595 | 17.4538 | 0.1686 |
| 2 | 0.1571 | 21.8036 | 0.1375 | 0.1514 | 21.4999 | 0.1383 |
| 3 | 0.1426 | 22.8552 | 0.1107 | 0.1416 | 22.5334 | 0.1075 |
| 4 | 0.1374 | 22.8948 | 0.1046 | 0.1374 | 23.0937 | 0.0927 |
| 5 | 0.1294 | 23.4292 | 0.0874 | 0.1304 | 23.4719 | 0.0899 |
| 6 | 0.1261 | 23.6784 | 0.0859 | 0.1274 | 23.7816 | 0.0945 |
| 7 | 0.1222 | 23.6329 | 0.0882 | 0.1264 | 23.6755 | 0.0844 |
| 8 | 0.1238 | 24.0451 | 0.0844 | 0.1217 | 23.9293 | 0.0803 |
| 9 | 0.1192 | 23.7569 | 0.0790 | 0.1186 | 23.9288 | 0.0781 |
| 10 | 0.1197 | 23.6757 | 0.0808 | 0.1194 | 23.3383 | 0.0897 |
| 11 | 0.1239 | 23.5051 | 0.0755 | 0.1203 | 24.0476 | 0.0803 |
| 12 | 0.1227 | 23.9581 | 0.0809 | 0.1208 | 23.9078 | 0.0853 |
| 13 | 0.1186 | 24.2616 | 0.0770 | 0.1189 | 23.9740 | 0.0799 |
| 14 | 0.1184 | 24.1779 | 0.0758 | 0.1178 | 24.2081 | 0.0737 |
| 15 | 0.1156 | 24.1974 | 0.0811 | 0.1175 | 24.5342 | 0.0793 |

*Note.* Ablation Study.

**Figure 8**

*Training Dynamics, PSNR Progression, and High-Frequency Loss Trajectories*

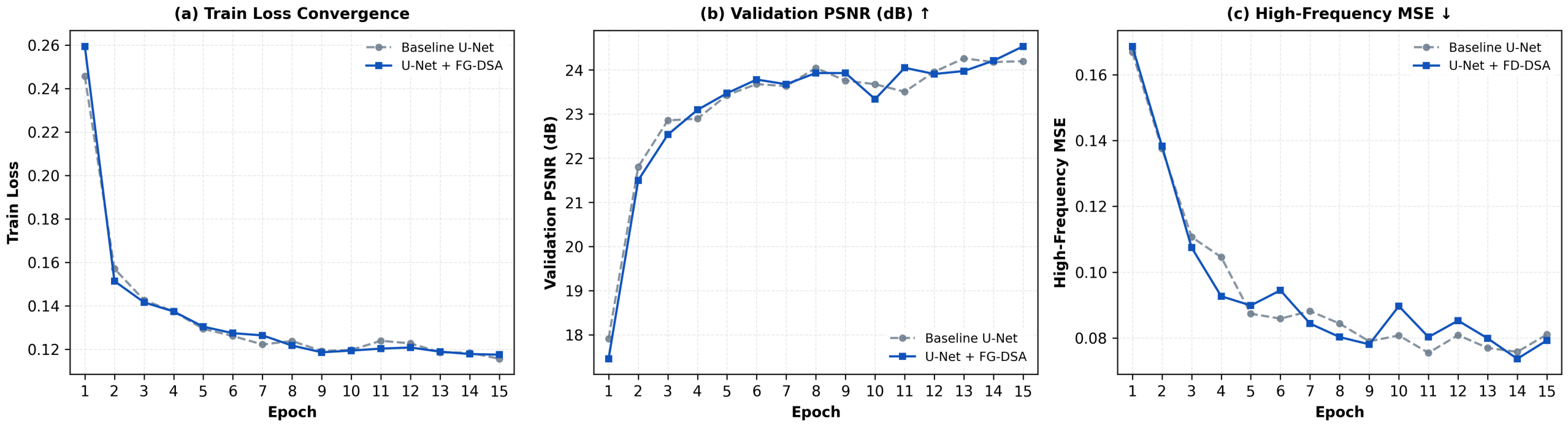

*Note.* Ablation Study.

The generative performance benchmark and metric improvement analysis (Table 7) shows that FG-DSA improves reconstruction quality and high-frequency detail preservation in the evaluated single-step image restoration task, delivering a 0.3368dB improvement in peak signal-to-noise ratio (PSNR) from 24.1974 to 24.5342 and a 2.2125% reduction in high-frequency error, with MSE from 0.0811 to 0.0793, over the baseline Generative U-Net. These results suggest that incorporating FG-DSA into the generative bottleneck can improve reconstruction of fine spatial details and image fidelity, although these gains come with a measurable increase in inference latency. The proposed model introduces a 6.08% increase in latency compared with the baseline (Table 8).

**Table 7**

*Generative Performance Benchmark and Metric Improvement Analysis*

| | PSNR | High-Frequency MSE |
|---|---|---|
| Baseline Generative UNet (No FG-DSA) | 24.1974 dB | 0.0811 |
| Proposed Generative UNet (With FG-DSA) | 24.5342 dB | 0.0793 |

*Note.* Ablation Study.

Furthermore, in order to quantify the computational cost of the improvements in high-frequency detail, we assess the model in terms of parameter count and single-image inference latency (Table 8). The baseline Generative U-Net has 0.3075M parameters and an average inference latency of 11.9858 ± 0.0431ms per image. FG-DSA increases the parameter count to 0.3913M, an absolute increase of 0.0838M (27.2687%). The average inference latency is now 12.7142 ± 0.0457ms per image, an additional 0.7284ms (6.0773%). Thus, FG-DSA incurs a non-trivial parameter overhead, though the absolute increase in inference latency is relatively small in this case.

**Table 8**

*Computational Efficiency and Structural Overhead Assessment*

| Model Architecture | Params (M) | Parameter Overhead | Latency (ms/img) | Latency Overhead |
|---|---|---|---|---|
| Baseline Generative UNet | 0.3075 | — | 11.9858 ± 0.0431 | — |
| Proposed Generative UNet (With FG-DSA) | 0.3913 | Increases in 0.0838 (Increases by 27.2687%) | 12.7142 ± 0.0457 | Increases in 0.7284 ms (Increases by 6.0773%) |

## VII. Conclusion

### (VII.1) Conclusion of Research Question 1

In Research Question 1, we investigated whether explicitly integrating a frequency-guided prior via the proposed FG-DSA module can counteract edge degradation, overcome the deep representation bottleneck, and enhance dense semantic feature representations in lightweight segmentation networks. Quantitative dynamics and qualitative evidence from multi-stage optimization trajectories, targeted spatial boundary protocols, bottleneck activation analyses, and full-resolution semantic segmentation benchmarks confirm the design and effectiveness of the FG-DSA module.

Empirical optimization trajectories confirm that FG-DSA restructures deep feature learning via selective filtering of spatial frequencies rather than uniform, smooth spatial convolutions. There is a very brief parameter warm-up period (Epochs 0-4) as the module learns to isolate the high-frequency spectral components, followed by a clear optimization crossover at Epoch 5. From that point on, it overcomes the capacity and spatial-smoothing constraints of the lightweight convolutional backbone and continues to reduce the loss to a floor of 0.4587, achieving a net loss reduction of 0.1505 over the baseline. This clearly demonstrates that the Fourier-guided high-pass gating provides superior capacity for dense feature learning and significant long-term optimization momentum.

Fine-grained structural evaluation shows that FG-DSA directly addresses and fixes spatial edge degradation in areas where other architectures simply cannot. We established a targeted ±5px morphological boundary evaluation for hard-class slender instances (Pole, SignSymbol, Fence, Pedestrian, Bicyclist), which effectively removed inflation in the background denominator and allowed us to isolate the true high-frequency gains. The module achieves a significant improvement in Hard-Class Boundary mIoU, up 10.1422% from 18.4731% to 28.6153%, and with sample-wise hard-class boundary pixel error reduction rates ranging from 11.6422% to 12.0574%. The qualitative error map distributions confirm that the corrected pixels are predominantly along complex building contours, sidewalk edges, and slender foreground objects, showing that FG-DSA successfully suppresses boundary over-smoothing and eliminates spatial edge blurring.

Deep bottleneck feature analysis reveals the mechanism behind these spatial enhancements. In a standard deep encoder, multi-scale down-sampling causes massive feature attenuation and over-smoothing along thin contours. FG-DSA achieves this by injecting Fourier high-pass spectral priors directly into the deep bottleneck, resulting in a 22.9595% amplification of the mean hard-class boundary feature response. This leads to high contrast and very sharp activation patterns before decoder upsampling. Spectral amplification manifests as continuous, contiguous semantic regions with well-preserved object boundaries, effectively suppressing fragmentation in misclassifications and the background. These results show that FG-DSA can improve spatial feature representation and improve boundary-related performance in semantic segmentation.

### (VII.2) Conclusion of Research Question 2

For Research Question 2, we explored whether FG-DSA could be integrated into a generative U-Net for single-step image restoration. We examined its convergence behavior, reconstruction quality, high-frequency spectral error, and computational overhead. The results show that FG-DSA improves reconstruction quality and helps retain high-frequency details, as reflected in higher PSNR and lower high-frequency MSE. These findings suggest that frequency-guided attention may also benefit generative image restoration. However, the evaluation is limited to the single-step restoration setting considered in this study. Further experiments are needed to determine whether similar benefits can be achieved in multi-step diffusion processes and other generative architectures.

Convergence and optimization dynamics show that FG-DSA dramatically improves the learning capacity and stability of the generative backbones. Over the 15-epoch training trajectory, the proposed Generative U-Net with FG-DSA clearly outperforms the baseline. They start off at roughly the same level in terms of reconstruction at Epoch 1, and the proposed model then rapidly surges ahead. It achieves its lowest HF-MSE of 0.0737 at Epoch 14 and a peak validation PSNR of 24.5342dB at Epoch 15. The clean optimization path demonstrates that introducing frequency-guided attention into the deep generative bottleneck allows the network to learn richer high-frequency structural content without compromising the stability of the end-to-end training.

Spatial detail and fidelity benchmarks confirm that FG-DSA successfully removes spatial blur and improves the richness of the fine synthesized textures. Compared to the baseline Generative U-Net, including FG-DSA resulted in a respectable +0.3368dB improvement in PSNR, from 24.1974dB to 24.5342dB, and a 2.2125% reduction in high-frequency MSE, from 0.0811 to 0.0793. The metrics clearly demonstrate that injecting Fourier high-pass spectral content at the deep feature level preserves the sharpness of structural contours and enhances local high-frequency detail that would otherwise be lost during generative feature upsampling.

The computational efficiency analysis highlights a trade-off between reconstruction quality and computational cost. Specifically, FG-DSA increases the model parameter count by 27.2687% (an additional 0.0838M parameters) and inference latency by 6.0773% (0.7284 ms per image). While the increase in inference time is relatively small under the current experimental setting, the additional parameter overhead remains noteworthy. This suggests that further optimization could be beneficial, particularly when deploying the framework in resource-constrained environments.

Overall, the results suggest that FG-DSA can improve reconstruction quality and high-frequency detail preservation in the evaluated single-step image restoration setting. Although these findings indicate its potential beyond discriminative tasks, further experiments are needed to determine whether similar benefits extend to other generative architectures and multi-step diffusion processes.

## VIII. Discussion

The main strength of FG-DSA lies in its design, which balances spatial detail preservation with computational efficiency in lightweight vision models. By combining a 2D-FFT high-pass filter with Depthwise Separable Attention, the module incorporates frequency-domain information into the spatial attention mechanism. While FG-DSA is intended to offer a more efficient alternative to computationally intensive attention mechanisms, its integration does introduce additional computational costs. In our generative experiments, the parameter count increases by 27.2687%, while single-image inference latency rises by 6.0773%. This suggests that the improvements in reconstruction quality come with a measurable computational trade-off. Although the increase in inference latency is relatively modest, the additional parameters remain an important consideration, particularly for applications with limited computational resources.

Our experiments also suggest that FG-DSA may be useful beyond discriminative tasks such as semantic segmentation, as it can be incorporated into generative image restoration architectures. However, this finding should be interpreted within the scope of the current evaluation, which focuses on single-step image restoration rather than a complete diffusion process. Although FG-DSA could potentially be integrated into diffusion-based architectures, its effectiveness in multi-step diffusion settings has not yet been established. Further experiments involving iterative denoising and diffusion sampling are therefore needed to determine whether the observed benefits extend to more complex generative processes.

Another consideration is that the actual inference speed of FG-DSA may depend on the underlying hardware. While the 2D-FFT provides a useful way to extract frequency-domain information, its implementation can require substantial memory bandwidth and may not be as well optimized as standard convolution operations on certain GPUs. As a result, the actual latency on edge devices could be higher than the computational estimates suggest. Future work could explore more hardware-friendly alternatives to global frequency filtering, including localized wavelet transforms and fixed spatial high-pass convolutions. Kernel fusion may also help reduce memory-access overhead and improve deployment efficiency on resource-constrained devices.

## IX. Implication & Contribution

The proposed FG-DSA framework is a major contribution in terms of both theory and practice for lightweight computer vision. It provides a fundamental solution to the classic trade-off between preserving fine-grained detail and computational efficiency that has long plagued the field. The approach marries frequency-domain prior injection with linear-complexity spatial attention. The key technical contributions and wider implications of the work are as follows.

- **Novel Architecture for Detail-Preserving Lightweight Vision:** The main technical contribution of this work is the introduction of FG-DSA. This new module is designed specifically for vision tasks with very limited resources and directly addresses spatial detail loss and boundary degradation caused by repeated feature downsampling.

- **Efficient Integration of Frequency Priors with Linear Attention:** We propose a novel hybrid mechanism that fuses a 2D-FFT high-pass filter with a depthwise separable attention module. It achieves this by explicitly injecting frequency-domain structural priors into spatial attention, allowing lightweight networks to capture sharp, fine-grained object boundaries while maintaining linear computational complexity.

- **Generalization Across Discriminative and Generative Tasks:** Another contribution of this work is the initial investigation of FG-DSA in both discriminative and generative settings. Our experiments show that the module can be incorporated into semantic segmentation and single-step generative image restoration architectures. However, this integration may add computational costs, including increases in model parameters and inference latency. Further work is needed to assess how well FG-DSA performs across different hardware platforms and whether its benefits extend to more complex generative architectures.

- **Bridging Spatial and Frequency Domains in Lightweight Model Design:** This study investigates how frequency-domain information can complement spatial attention mechanisms in lightweight vision architectures. By introducing high-frequency guidance into feature processing, FG-DSA helps preserve structural boundaries and fine-grained details more effectively. The experimental results provide preliminary evidence that combining frequency-domain information with lightweight spatial operations can be beneficial. However, further studies are needed to assess the computational trade-offs of this approach and examine whether its benefits can be consistently maintained across different tasks and architectures.

- **Exploring Unified Perception and Generation Architectures:** This work examines the potential of FG-DSA for both semantic segmentation and single-step generative image restoration. The experimental results suggest that frequency-guided attention can be integrated into different types of vision architectures, indicating its potential beyond conventional discriminative tasks. However, further evaluation is needed to determine how well the approach extends to more complex generative models, particularly multi-step diffusion architectures. Its computational efficiency and practical applicability across different hardware platforms also warrant further investigation.

In summary, this work presents a unified and highly practical paradigm for overcoming the long-standing efficiency-accuracy bottleneck in lightweight visual computing. FG-DSA demonstrates how frequency-domain high-pass filtering can be combined with linear-complexity spatial attention to improve fine-grained feature representation, while introducing measurable parameter and computational overhead. Beyond its contributions to dense prediction and single-step generative image restoration, this work provides an initial exploration of frequency-guided attention in different visual learning settings. FG-DSA highlights the importance of enabling lightweight vision systems to preserve the fine-grained visual details that are essential for meaningful visual understanding.

## XI. Appendix

**Python Code Availability:** The source code, architecture implementations, and evaluation notebook (FG-DSA YIYAO YANG.ipynb) for FG-DSA are stored on GitHub in a private repository (under the CC BY-NC-ND license): https://github.com/yiyaoyang2026/FG-DSA-Frequency-Guided-Depthwise-Separable-Attention-for-Efficient-Dense-Prediction-. To protect intellectual property, the code is available from the author upon reasonable request.